\pdfoutput=1

\documentclass[11pt]{article}

\usepackage[]{ACL2023}

\usepackage{times}
\usepackage{latexsym}

\usepackage{multirow}
\usepackage{amsmath}
\usepackage{svg}
\usepackage[T1]{fontenc}

\usepackage{tikz}
\usepackage{pgfplots}
\usetikzlibrary{patterns}
\usepackage{collcell}
\usepackage{times}
\usepackage{latexsym}

\usepackage{multirow}
\usepackage{amsmath}
\usepackage{svg}
\usepackage[T1]{fontenc}
\usepackage{multirow}

\usepackage{microtype}

\usepackage[utf8]{inputenc}

\DeclareUnicodeCharacter{1E41}{\d{m}}
\DeclareUnicodeCharacter{0304}{\={}}
\DeclareUnicodeCharacter{0325}{\d{}}
\usepackage{microtype}

\usepackage{inconsolata}

\title{Unsupervised Approach to Evaluate Sentence-Level Fluency: \\Do We Really Need Reference?}

\author{Gopichand Kanumolu\thanks{* Authors contributed equally} ,
  Lokesh Madasu\footnotemark[1] , 
  Pavan Baswani\footnotemark[1] ,
  Ananya Mukherjee\footnotemark[1] , \\
  \textbf{Manish Shrivastava}\\
  Language Technologies Research Center, KCIS, IIIT Hyderabad, India.\\
\texttt{\{gopichand.kanumolu, lokesh.madasu\}@research.iiit.ac.in} \\
\texttt{\{pavan.baswani, ananya.mukherjee\}@research.iiit.ac.in} \\
\texttt{m.shrivastava@iiit.ac.in}
}

\begin{document}
\maketitle
\begin{abstract}
Fluency is a crucial goal of all Natural Language Generation (NLG) systems. Widely used automatic evaluation metrics fall short in capturing the fluency of machine-generated text. Assessing the fluency of NLG systems poses a challenge since these models are not limited to simply reusing words from the input but may also generate abstractions. Existing reference-based fluency evaluations, such as word overlap measures, often exhibit weak correlations with human judgments. This paper adapts an existing unsupervised technique for measuring text fluency without the need for any reference. Our approach leverages various word embeddings and trains language models using Recurrent Neural Network (RNN) architectures. We also experiment with other available multilingual Language Models (LMs). To assess the performance of the models, we conduct a comparative analysis across \textit{10 Indic languages}, correlating the obtained fluency scores with human judgments. Our code and human-annotated benchmark test-set for fluency is available at \url{https://github.com/AnanyaCoder/TextFluencyForIndicLanaguges}.
\end{abstract}

\section{Introduction}

Fluency measures the quality of the generated text output from the model without considering the reference text \citep{fluencybook}. It accounts for grammar, spelling, word choice, and style characteristics. As stated by \citealp{martindale-carpuat-2018-fluency}, maintaining text fluency avoids misapprehensions, makes interactions more realistic,  and leads to higher user satisfaction and trust. Thus, fluency evaluation is essential for developing better models or screening unacceptable generations. 

Measuring fluency is important for evaluating the performance of NLG tasks like summarization, paraphrase generation, image captioning, etc. 
Fluency evaluation can be done by humans or automated metrics. For manual evaluation, proficiency in the target language is necessary. Besides, it is time-consuming, expensive, and requires a lot of human effort. 
\begin{table}[]
\centering
\begin{tabular}{ll}
\hline
\textbf{Fluency Scale} & \textbf{Example} \\ \hline
\begin{tabular}[c]{@{}l@{}}Perfect \\ Fluency\end{tabular} & \begin{tabular}[c]{@{}l@{}}You must include exercise\\in your daily routine.\end{tabular} \\ \hline
\begin{tabular}[c]{@{}l@{}}Acceptable \\ Fluency\end{tabular} & \begin{tabular}[c]{@{}l@{}}Delighted with the \textcolor{red}{respsons}\\ from the crowd.\end{tabular} \\ \hline
\begin{tabular}[c]{@{}l@{}}Low Quality \\ Fluency\end{tabular} & \begin{tabular}[c]{@{}l@{}}Government clears \underline{...} related\\to land acquisition \underline{...} the\\ \textcolor{red}{irrigaton} project.\end{tabular} \\ \hline
Incomprehensible & \begin{tabular}[c]{@{}l@{}}Six after dead construction\\in collapse wall.\end{tabular} \\ \hline
\end{tabular}
\caption{Examples of classifying sentences as per fluency scale. Misspelt words are highlighted in red and missing words are denoted by \underline{...}.}
\label{tab:my-table1}
\end{table}

To the best of our knowledge, there are no specific automatic evaluation metrics to measure text fluency without reference text. However, researchers use lexical overlap metrics to evaluate text quality in terms of fluency with the help of reference text \cite{lin-och-2004-automatic,papineni-etal-2002-bleu}. As a result, fluency is often manually assessed, which is expensive, laborious, and irreproducible.

Fluency evaluation of sentences has been a linguistic ability of humans and has been an arguable subject for many decades in linguistics, psychology, and cognitive science. The question has been raised whether the grammatical knowledge underlying this ability is probabilistic or categorical \cite{Chomsky+2020,ManningManuscript-MANPS,Sprouse2007ContinuousAC}. In a similar context, \citealt{https://doi.org/10.1111/cogs.12414} have illustrated that neural language models (LM) can be used to model human acceptability judgments. \citealt{https://doi.org/10.48550/arxiv.1809.08731} proposed \textbf{S}yntactic \textbf{L}og-\textbf{O}dds \textbf{R}atio (\textbf{SLOR}) score, which leverages \textit{sentence log probability}, normalized by \textit{unigram probability} and \textit{sentence length}, to correlate well with human ratings at the sentence level. They investigated the practical implications of \citealt{https://doi.org/10.1111/cogs.12414}'s findings for fluency evaluation of NLG, using the task of automatic compression. They also introduced a) WPSLOR: a Word-Piece \cite{https://doi.org/10.48550/arxiv.1609.08144}-based version of SLOR and b) ROUGE-LM: a combination of WPSLOR and ROUGE \cite{lin-och-2004-automatic}, the latter being a reference-based fluency evaluation approach. We extend their work by applying the syntactic log odds ratio to the 6 Indo-Aryan\footnote{Bengali (bn), Gujarati (gu), Hindi (hi), Marathi (mr), Odia (od), Sinhala (si)} and 4 Dravidian languages\footnote{Kannada (ka), Malayalam (ml), Telugu (te), Tamil (ta)}. Our main motivation is to investigate this approach for morphologically rich languages. 

Data scarcity is a very common problem faced by the languages in the Indian subcontinent \cite{https://doi.org/10.48550/arxiv.2004.09095}. In addition, the quality of the available datasets is highly questionable. 

Existing monolingual corpora have several issues like presence of non-unique sentences, junk/unwanted characters; requires additional cleaning and hence there is an overhead of pre-processing and de-noising. 
Therefore we retrieve clean, filtered data from regional news websites (see Appendix Table \ref{tab:crawling_sources}) and train multiple LMs using RNN \cite{hochreiter1997long, cho2014properties} and transformer architectures \cite{devlin-etal-2019-bert} leveraging various embedding techniques \cite{bojanowski2017enriching, heinzerling-strube-2018-bpemb, kakwani2020indicnlpsuite, khanuja2021muril} and compute sentence-level fluency score with syntactic log odds ratio. To assess the quality of the models, it is necessary to compare the scores with humans. However, due to the non-availability of benchmark dataset exclusively for fluency evaluation of the Indian subcontinent languages; we create a corpora for each language along with the designated fluency scores by humans. Proficient native speakers assign these manual scores by following strict guidelines. Table \ref{tab:my-table1} depicts our fluency scale with example sentences. 
Ultimately we compute the Pearson Product Moment Correlation score of the calculated fluency scores with the human assessments.

Our contributions in this work are in two-fold:

\begin{itemize}
    \item We release 5K human annotated sentences (500 sentences for each language) which can be further used as a benchmark test-set for fluency evaluation.
    \item  We present our reference-free, unsupervised experiments to measure fluency for 6 Indo-Aryan and 4 Dravidian languages;
\end{itemize}

\section{Existing Approaches}
\label{sec:related-word}
Research in automatic evaluation of NLG systems has increasingly become important. However, the task of measuring \textit{text fluency} remain barely investigated. NLG researchers often use
word-overlap based methods to measure text fluency for tasks such as text compression, machine translation, text generation, etc. Existing n-gram overlap metrics like BLEU \cite{papineni-etal-2002-bleu} and ROUGE \cite{lin-och-2004-automatic} use reference text to evaluate text quality in terms of fluency.

\subsection{Can Readability Scores Measure Fluency?}

As text fluency also depends on its vocabulary and sentence structure, we can make an assumption that readability scores can act as a tool to measure fluency. Readability refers to how easy it is to read and understand a text \cite{dubay2004principles}. 
For many readability measures, the number of syllables is one of the building blocks on which the formula is created. Generally,  readability formulas use syllable counting to judge how easy or hard a piece of text is to read.
Therefore, any inaccuracies or inconsistencies in counting syllables will have consequences for the accuracy of the readability measurement. In the English readability formula, word length, often measured by the number of syllables, is the key in how difficult or easy a text is to read. In formulas such as Flesch Reading Ease \cite{Kincaid1975DerivationON} and Gunning Fog Index \cite{gunning}, three syllables or more words are categorized as ‘hard’ words. We cannot apply this in the context of Indian subcontinent languages. In a study of text readability in Bengali, researchers state that in Bengali, polysyllabic words are common in everyday use \cite{article_bn}. For example, "Āmādēra" means "ours" in Bengali, which has three syllables and is not a difficult word, but for English readability formula this would count as a complex word. Most of the commonly used words in Indian subcontinent languages are polysyllabic.
Hence syllables, both in terms of how to count them and how important they are when you count them, pose a challenge to the development of readability formula for languages of Indian subcontinent.

\subsection{LM based Approaches}
It is widely believed that much of human cognition is probabilistic \cite{article_ALLAN}. It is also a well known fact that a language model is a probabilistic model that assigns probabilities to words and sentences. Therefore, we can safely use Language Model (LM) as a tool to measure fluency. LMs assign higher probabilities to sentences that are syntactically correct and lower probabilities to sentences that are not. \citealt{https://doi.org/10.48550/arxiv.1809.08731} introduced LM-based approaches (SLOR, WPSLOR, ROUGE-LM) to compute the fluency score by normalizing sentence probability with unigram probability and sentence length. They proved that ROUGE-LM showcased better correlation with humans. However, ROUGE-LM is a reference-based approach that demands gold references. Finding reliable references for evaluating the fluency of various NLG tasks is not always possible.

Thus, our research is inclined towards \textit{unsupervised} and \textit{reference-free evaluation} of text fluency. 
Here, we compute fluency using several combinations of LMs trained using various embeddings. Our work mainly follows the path started by \citealt{https://doi.org/10.48550/arxiv.1809.08731} and performs intensive experiments on 6 Indo-Aryan and 4 Dravidian languages to determine fluency at a sentence level. 


\section{Reference-Free Sentence-level Fluency Evaluation}
\label{sec:methodology}

In this work, we make an attempt to extend the existing reference-free approach by applying to various morphologically rich languages to get the sentence-level fluency score from the trained language model.

\begin{figure}[h]
    \centering
    \includegraphics[scale=0.7]{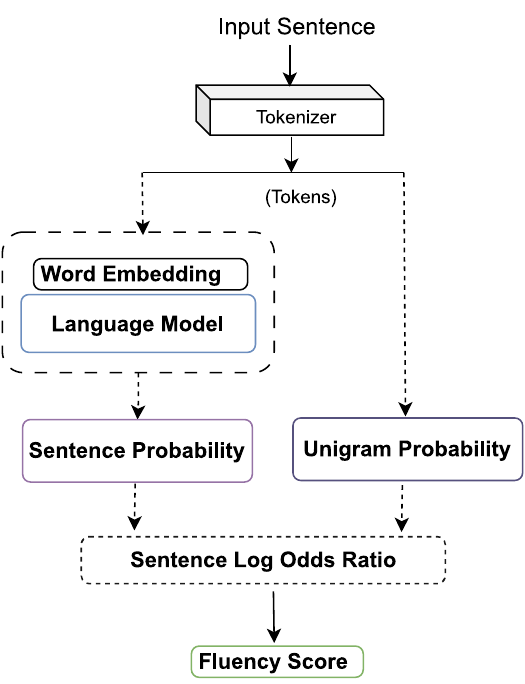}
    \caption{Illustration of our model architecture.}
    \label{fig:Architecture}
\end{figure}

Figure \ref{fig:Architecture} illustrates our approach in detail. Initially, an input sentence is tokenized, and each token is fed to the embedding layer. Using these embeddings, the language model assigns a probability to the given sentence. On the other hand, unigram probability is computed using a list of tokens. To determine sentence fluency, we compute the Syntactic Log-Odds Ratio (SLOR) score by subtracting unigram probability from the sentence probability and further normalizing it by the sentence length (refer Eqn. \ref{eqn:slor_score}).

\subsection{Language Models}
Language Modeling (LM) is the task of predicting what word will come next, or, more broadly, a system that computes the likelihood of a sentence (word sequence) or the probability of the next word in a text sequence. The simplest language model is the N-gram, and its performance is restricted by its simplicity and incapable of achieving fluency, sufficient linguistic variation, and proper writing style for extended documents. For these reasons, despite their complexity, neural networks (NN) are being considered as the next primary standard. Recurrent Neural Networks (RNN) have become a base architecture for any sequence. 
RNNs are currently regarded as the usual text architecture, but they have their own issues: they cannot recall prior content for lengthy periods of time and struggle to construct long relevant text sequences due to exploding or disappearing gradient concerns. 

As a result, new architectures, such as Long Short Term Memory (LSTM) \cite{hochreiter1997long} and Gated Recurrent Units (GRU) \cite{cho2014properties}, were created and became the state-of-the-art solution for many language generation problems.

\subsection{Text Representation}
In this subsection, we discuss the various text representation techniques used in our experiments. 

\subsubsection{FastText Embeddings}
FastText \cite{bojanowski2017enriching} is an extended version of the \citealt{Mikolov}'s embedding. It is based on skip-gram model and uses subword information to generate the embeddings for a given word. Instead of learning vectors for words directly, FastText represents each word as an n-gram of characters, which aids it to provide embeddings for rare words.

\subsubsection{Byte-Pair Embeddings (BPEmb)}
BPEmb \cite{heinzerling-strube-2018-bpemb} is pre-trained on Wikipedia using the Byte-Pair Encoding technique. Due to the advantage of being subword embeddings, it has the ability to generate embeddings for out-of-vocabulary words. 

\subsubsection{IndicBERT}

IndicBERT \cite{kakwani2020indicnlpsuite} is a multilingual ALBERT \cite{lan2019albert} model trained on large-scale corpora, covering 12 Indian languages. It is pre-trained on AI4Bharat's monolingual corpus containing 8.9B tokens.


\subsubsection{MuRIL}
MuRIL (Multilingual Representations for Indian Languages) \cite{khanuja2021muril} is a BERT based multilingual language model especially built for 16 Indian languages that is trained using large text corpora of Indian languages.

\subsection{Fluency Measure with LM}

\subsubsection{Syntactic Log-Odds Ratio (SLOR)}
SLOR \cite{https://doi.org/10.48550/arxiv.1809.08731} assigns a score to a given sentence that is computed using the log-probability
of the sentence given by the language model, normalized by its unigram log-probability and length. Sentence SLOR score can be computed using the Equation \ref{eqn:slor_score}.
\begin{equation}
    \begin{split}
    SLOR(S) & = \frac{1}{|S|} (ln (PM(S)) - ln(p_{u}(S)))\\
    p_{u}(S) & = \prod_{t \in S} p(t)
    \end{split}
    \label{eqn:slor_score}
\end{equation}
Where, |S| is the length of the sentence in terms of tokens; PM(S) is the sentence probability returned by the model; $P_u(S)$ is the unigram probability; p(t) denotes the unconditional probability of a token `t` given no context.

\subsubsection{WordPiece SLOR (WPSLOR)}
WPSLOR is the modified SLOR score computed by considering the word pieces instead of words. WordPieces produce a smaller vocabulary, which reduces model size and training time while improvising the handling of rare words by partitioning them into more frequent segments. However, they contain more information than characters.

\section{Dataset Description}
\label{sec:dataset}
\subsection{Data Collection}

\begin{figure}[!h]
    \centering
    \includegraphics[scale=0.5]{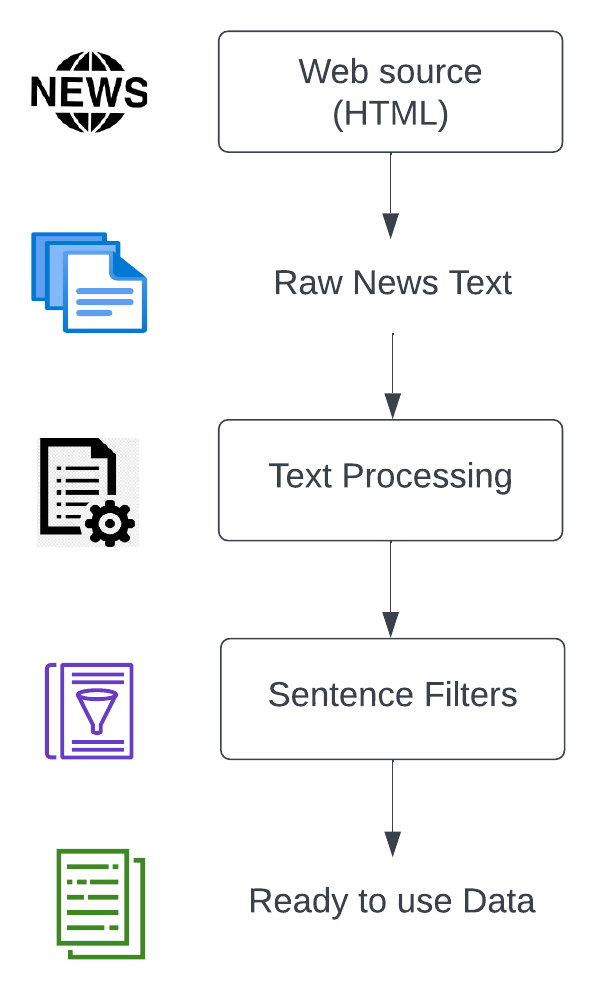}
    \caption{Data collection pipeline}
    \label{fig_dataCollection}
\end{figure}

We mined text from web sources for 10 regional news websites (refer Appendix Table. \ref{tab:crawling_sources}) using the Python libraries: Requests\footnote{\url{https://pypi.org/project/requests/}}, BeautifulSoup\footnote{\url{https://pypi.org/project/beautifulsoup4/}} \cite{Zheng2015ASO} and Selenium\footnote{\url{https://pypi.org/project/selenium/}}. Figure \ref{fig_dataCollection} depicts the data collection pipeline adapted in our work. This data collection is performed in a controlled setting to filter unwanted ads, HTML tags, URLs and other embedded social media content by maintaining a site-specific scraping script. These scripts extract only the news text from sources and avoid most of the noisy data as a primary filter. Later the raw text is cleaned and processed, resulting in a total of 1.025M samples\footnote{10 $*$ (100K $+$ 1K $+$ 1K $+$ 500) = 10,25,000}. From the processed and filtered samples, for each language, we distributed 100K samples for training, 1K for validation, 1K for testing. In addition, for the language-wise fluency test set, we translated 500 English samples for human annotations\footnote{500 * 10 Languages = 5000} (see Section \ref{sec:data_annotation}).


\begin{table*}[]
\centering
\resizebox{\textwidth}{!}{
\begin{tabular}{cll}
\hline
\textbf{Score} & \textbf{Fluency Quality} & \textbf{Guideline} \\ \hline
\begin{tabular}[c]{@{}l@{}}3\end{tabular} & Perfect & \begin{tabular}[c]{@{}l@{}}Perfectly fluent sentence without any syntactic or grammatical error.\\
\textbf{Example}: Government clears issues related to land \\ acquisition of the irrigation project.\end{tabular} \\ \hline

\begin{tabular}[c]{@{}l@{}}2\end{tabular} & Moderate & \begin{tabular}[c]{@{}l@{}} 
Sentence is predominantly fluent but contains either \\ a) misspelt word or \\ b) missing word or \\ c) multiple occurrence of a word.\\
\textbf{Example}: Delighted with the \textcolor{red}{repsonse} from the crowd.\end{tabular} \\
\hline
\begin{tabular}[c]{@{}l@{}}1\end{tabular} &  Low Quality & \begin{tabular}[c]{@{}l@{}}Partially fluent sentence: \\ a) only half of the sentence is fluent or \\ b) more than 1 missing words or\\ c) more than 1 misspelt words or \\ d) contains individual fluent word-groups with missing coherence \\between them.\\
\textbf{Example}: Government clears \underline{...} related to land acquisition \underline{...} the \textcolor{red}{irrigatoin}\\project.

\end{tabular} \\ \hline
\begin{tabular}[c]{@{}l@{}}0\end{tabular} & Incomprehensible & \begin{tabular}[c]{@{}l@{}} Inarticulate/ non-fluent sentence\\
\textbf{Example}: Six after dead construction in collapse wall.
\end{tabular} \\

\end{tabular}
}
\caption{Annotation guidelines with fluency scores and examples. Misspelt words are highlighted in red and
missing words are denoted by \underline{...}}
\label{tab:guidelines}
\end{table*}

\subsubsection{Preprocessing}

Data pre-processing is crucial in building any machine learning (ML) model. We build custom pre-processors to extract the sentences from the articles. During this process, we clean the sentences by removing special symbols, junk characters, etc. The additional \textit{spaces, tab-spaces and new-lines} are replaced with a single space. The pre-processed sentences are filtered based on the number of tokens at the sentence level. It was also taken care that the sentences do not contain any English text. We consider the sentences having tokens\footnote{used space as a delimiter} in the range of 8-25 (to avoid too short/long sentences).

\subsection{Data Annotation}
\label{sec:data_annotation}
We collect the human annotated data with the help of proficient native speakers who are computer science graduates with a background in the field of NLP. The annotators were provided with guidelines and were asked to assign a fluency score to each sentence.

\subsubsection{Need for Data Annotation}

To validate our experiments, we require a standard human annotated test set with fluency scores for each sentence. To the best of our knowledge, there are no human annotated datasets available for measuring fluency of text generated by NLG systems.

\subsubsection{Annotation Guidelines}

There are no annotation guidelines on assigning a fluency score to a sentence. After careful observation of outputs generated by NLG system, we identified spelling mistakes, redundant words, and coherence issues as common mistakes. 
With these observations, we penalize the fluency score for the sentence which contains most of the above-mentioned issues. Table \ref{tab:guidelines} details the proposed annotation guidelines for fluency scale.

\subsubsection{Statistics of Annotated Data}

Native speakers from 10 languages have voluntarily participated in the annotation exercise. Each annotator was made familiar with the guidelines and the process was clearly mentioned. Each individual was assigned with a total of 500 sentences \footnote{The sentences are translated using Google Translate (\url{https://translate.google.com/})} to post-edit and induce errors based on the guidelines. To have a balanced distribution, the annotators were instructed to induce errors in such a manner that the ultimate submission contained 200 sentences with a fluency score of 3 and 100 sentences each with fluency scores of 2, 1, and 0. Therefore, we obtain 40\% fluent and 60\% non-fluent sentences.
\section{Experiments and Discussion}
\label{sec:experiments}
In this section, we describe the various implementation details and discuss the models' performance.

\subsection{Model Implementation}

Figure~\ref{fig:Architecture}, provides the overview of the proposed methodology to measure text fluency. We train an LM on scraped news articles from various news websites (see Appendix Table~\ref{tab:crawling_sources} ). To train the LM, we use the three variants of the Recurrent Neural Network (RNN) architecture such as LSTM, GRU and Bi-LSTM with the four different embedding techniques: FastText, BPE, IndicBERT and MuRIL embeddings. All the models are trained with 1 GPU support. Each RNN variant has 2 layers with 512 hidden activation units in each layer and a sequence of 128. To tune the LM parameters, we employ Categorical Cross Entropy as the loss function and Adam optimizer \cite{https://doi.org/10.48550/arxiv.1412.6980} with a learning rate of 0.001. We adopt early stopping criteria based on the validation loss as a regularization step to avoid over-fitting.
\subsection{Zero-shot and Fine-tuned Experiments}
We additionally experimented using existing pre-trained multilingual BERT LM\footnote{\url{https://huggingface.co/bert-base-multilingual-uncased}. Not available for Odia and Sinhala} \cite{devlin-etal-2019-bert} and Muril LM\footnote{\url{https://huggingface.co/google/muril-base-cased}. Not available for Sinhala} \cite{khanuja2021muril}.

Using these models, we performed zero-shot inferences and also fine-tuned them. The pre-trained model is fine-tuned on the same training data for 5 epochs with a learning rate of $2e^{-5}$, batch size of 8 and AdamW optimizer \cite{https://doi.org/10.48550/arxiv.1711.05101}.

\subsection{Results and Analysis}
Using these various trained and fine-tuned models, we independently compute fluency by calculating the SLOR score. To evaluate the models, we compute the Pearson product-moment correlation $(C)$ \citep{Benesty2009} between fluency scores $(F)$ and human ratings $(H)$ using Equation \ref{eqn:corr}, where, $N$ is the total number of samples (500 sentences). Pearson correlation estimates the degree of statistical relationship. Consequently, it can be logically reasoned that if a model has a high correlation with human judgments, it implies that the model correlates well with humans.
\begin{equation}
\small
\label{eqn:corr}
C =\frac{(N\sum\limits_{i=1}^N H\textsubscript{i}F\textsubscript{i}-(\sum\limits_{i=1}^N H\textsubscript{i})(\sum\limits_{i=1}^N  F\textsubscript{i}))}{\sqrt{N\sum\limits_{i=1}^N H\textsubscript{i}^2-(\sum\limits_{i=1}^N H\textsubscript{i}})^2\sqrt{N\sum\limits_{i=1}^N F\textsubscript{i}^2-(\sum\limits_{i=1}^N F\textsubscript{i})^2}}
\end{equation}

\begin{table*}[!ht]
\centering
\resizebox{\textwidth}{!}{
\begin{tabular}{l|lll|lll|lll|lll}

\hline
E & \multicolumn{3}{c|}{FastText} & \multicolumn{3}{c|}{BPEmb} & \multicolumn{3}{c|}{IndicBERT} & \multicolumn{3}{c}{MuRIL} \\ \hline \hline
Ln & \multicolumn{1}{l|}{Bi-LSTM} & \multicolumn{1}{l|}{LSTM} & GRU & \multicolumn{1}{l|}{Bi-LSTM} & \multicolumn{1}{l|}{LSTM} & GRU & \multicolumn{1}{l|}{Bi-LSTM} & \multicolumn{1}{l|}{LSTM} & GRU & \multicolumn{1}{l|}{Bi-LSTM} & \multicolumn{1}{l|}{LSTM} & GRU \\ \hline \hline
te & \multicolumn{1}{l|}{0.317} & \multicolumn{1}{l|}{0.355} & 0.313 & \multicolumn{1}{l|}{0.403} & \multicolumn{1}{l|}{0.410} & 0.327 & \multicolumn{1}{l|}{0.307} & \multicolumn{1}{l|}{0.290} & 0.270 & \multicolumn{1}{l|}{0.400} & \multicolumn{1}{l|}{\textbf{0.430}} & 0.280 \\
ml & \multicolumn{1}{l|}{0.119} & \multicolumn{1}{l|}{0.123} & 0.089 & \multicolumn{1}{l|}{0.328} & \multicolumn{1}{l|}{0.279} & 0.213 & \multicolumn{1}{l|}{0.145} & \multicolumn{1}{l|}{0.182} & 0.103 & \multicolumn{1}{l|}{0.300} & \multicolumn{1}{l|}{\textbf{0.300}} & 0.170 \\
ta & \multicolumn{1}{l|}{0.152} & \multicolumn{1}{l|}{0.124} & 0.134 & \multicolumn{1}{l|}{0.085} & \multicolumn{1}{l|}{0.130} & 0.098 & \multicolumn{1}{l|}{0.039} & \multicolumn{1}{l|}{0.019} & 0.020 & \multicolumn{1}{l|}{0.110} & \multicolumn{1}{l|}{\textbf{0.140}} & 0.080 \\
kn & \multicolumn{1}{l|}{0.201} & \multicolumn{1}{l|}{0.145} & 0.214 & \multicolumn{1}{l|}{0.381} & \multicolumn{1}{l|}{0.357} & 0.264 & \multicolumn{1}{l|}{0.398} & \multicolumn{1}{l|}{0.410} & 0.343 & \multicolumn{1}{l|}{\textbf{0.550}} & \multicolumn{1}{l|}{0.520} & 0.450
\\ \hline \hline
\end{tabular}
}

\caption{Pearson correlation scores of RNN-based LMs with human judgements for Dravidian Languages. \\E: Embeddings,     Ln: Language}
\label{tab:corr_dravidian}
\end{table*}

\begin{table*}[!ht]
\centering
\resizebox{\textwidth}{!}{

\begin{tabular}{l|lll|lll|lll|lll}
\hline
E & \multicolumn{3}{c|}{FastText} & \multicolumn{3}{c|}{BPEmb} & \multicolumn{3}{c|}{IndicBERT} & \multicolumn{3}{c}{MuRIL} \\ 
\hline \hline
Ln & \multicolumn{1}{l|}{Bi-LSTM} & \multicolumn{1}{l|}{LSTM} & GRU & \multicolumn{1}{l|}{Bi-LSTM} & \multicolumn{1}{l|}{LSTM} & GRU & \multicolumn{1}{l|}{Bi-LSTM} & \multicolumn{1}{l|}{LSTM} & GRU & \multicolumn{1}{l|}{Bi-LSTM} & \multicolumn{1}{l|}{LSTM} & GRU \\ \hline \hline
bn & \multicolumn{1}{l|}{0.191} & \multicolumn{1}{l|}{0.202} & 0.164 & \multicolumn{1}{l|}{0.335} & \multicolumn{1}{l|}{0.284} & 0.296 & \multicolumn{1}{l|}{0.171} & \multicolumn{1}{l|}{0.176} & 0.168 & \multicolumn{1}{l|}{0.308} & \multicolumn{1}{l|}{\textbf{0.350}} & 0.260 \\
od & \multicolumn{1}{l|}{0.030} & \multicolumn{1}{l|}{0.070} & -0.010 & \multicolumn{1}{l|}{0.190} & \multicolumn{1}{l|}{0.234} & 0.240 & \multicolumn{1}{l|}{0.217} & \multicolumn{1}{l|}{0.230} & 0.234 & \multicolumn{1}{l|}{0.200} & \multicolumn{1}{l|}{\textbf{0.200}} & 0.140 \\
hi & \multicolumn{1}{l|}{0.188} & \multicolumn{1}{l|}{0.216} & 0.126 & \multicolumn{1}{l|}{0.459} & \multicolumn{1}{l|}{0.456} & 0.438 & \multicolumn{1}{l|}{0.481} & \multicolumn{1}{l|}{0.488} & 0.478 & \multicolumn{1}{l|}{0.579} & \multicolumn{1}{l|}{\textbf{0.600}} & 0.560 \\
gu & \multicolumn{1}{l|}{0.377} & \multicolumn{1}{l|}{0.381} & 0.305 & \multicolumn{1}{l|}{0.432} & \multicolumn{1}{l|}{0.371} & 0.321 & \multicolumn{1}{l|}{0.282} & \multicolumn{1}{l|}{0.286} & 0.221 & \multicolumn{1}{l|}{0.474} & \multicolumn{1}{l|}{\textbf{0.500}} & 0.340 \\
mr & \multicolumn{1}{l|}{-0.022} & \multicolumn{1}{l|}{0.008} & 0 & \multicolumn{1}{l|}{0.403} & \multicolumn{1}{l|}{0.374} & 0.325 & \multicolumn{1}{l|}{0.515} & \multicolumn{1}{l|}{0.484} & 0.407 & \multicolumn{1}{l|}{0.500} & \multicolumn{1}{l|}{\textbf{0.500}} & 0.450 \\
si & \multicolumn{1}{l|}{0.359} & \multicolumn{1}{l|}{0.348} & -0.110 & \multicolumn{1}{l|}{0.411} & \multicolumn{1}{l|}{0.422} & 0.407 & \multicolumn{1}{l|}{-} & \multicolumn{1}{l|}{-} & - & \multicolumn{1}{l|}{-} & \multicolumn{1}{l|}{-} & - \\
\hline
\hline
\end{tabular}
}

\caption{Pearson correlation scores of RNN-based LMs with human judgements for Indo-Aryan Languages \\E: Embeddings,     Ln: Language}
\label{tab:corr_IndoAryan}
\end{table*}

\begin {figure*}[!h]
\centering
\begin{tikzpicture}
  \begin{axis}[  
    ybar, 
    bar width=4pt,
    enlargelimits=0.08,
    legend style={at={(0.5,-0.2)}, 
      anchor=north,legend columns=2},     
    ylabel={Pearson Correlation}, 
    symbolic x coords={te, kn, ta, ml, hi, mr, gu,  bn},  
    xtick=data,  
    width=.9\textwidth,
    height=.3\textheight    ]  
\addplot[pattern = dots] coordinates {(te,0.017) (kn,-0.011) (ta,0.019) (ml,0.007) (hi,0.129) (mr,0.054) (gu,0.061) (bn,0.011)};
\addlegendentry{mBERT Zero-shot}
\addplot[pattern = horizontal lines, pattern color = yellow] coordinates {(te,0.155) (kn,0.295) (ta,0.148) (ml,0.073) (hi,0.276) (mr,0.405) (gu,0.317) (bn,0.165) }; 
\addlegendentry{mBERT w/ Fine-tuning}
\addplot[pattern = vertical lines] coordinates {(te,-0.085) (kn,0.044) (ta,-0.049) (ml,-0.013) (hi,-0.06) (mr,-0.066) (gu,0.003) (bn,0.05)}; 
\addlegendentry{MuRIL Zero-shot}
\addplot[pattern = bricks, pattern color = red] coordinates {(te,0.44) (kn,0.5) (ta,0.23) (ml,0.37) (hi,0.4) (mr,0.5) (gu,0.55) (bn,0.4)}; 
\addlegendentry{MuRIL w/ Fine-tuning}
\addplot[pattern = crosshatch, pattern color = green] coordinates {(te,0.43) (kn,0.52) (ta,0.14) (ml,0.3) (hi,0.6) (mr,0.5) (gu,0.5) (bn,0.35) }; 
\addlegendentry{MuRIL+LSTM}

  \end{axis}
\end{tikzpicture}
\caption{Comparative analysis of correlation scores of various models. (The scores of Oriya and Sinhala are not reported)}
\label{fig:allmodels-pearson-correlation}
\end{figure*}
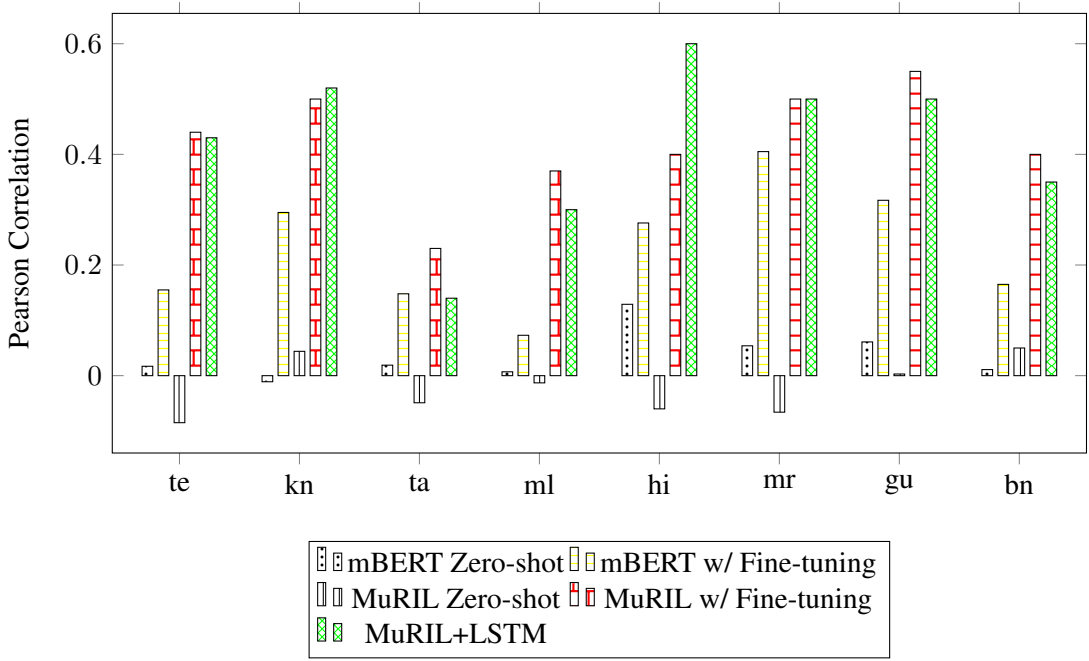

The correlation of fluency scores of RNN-based LMs with human judgments is reported in Table \ref{tab:corr_dravidian} and Table \ref{tab:corr_IndoAryan}. 
The findings from the reported tables indicate that the Muril+LSTM model performed the best, followed by RNNs trained using BPEmb. This is supported by the language-wise comparison of correlation scores with humans shown in Figure \ref{fig:allmodels-pearson-correlation}, in which we compare our best-trained model with zero-shot inferences and fine-tuned model. Both the \textit{Muril+LSTM} model and the \textit{Muril model with fine-tuning} demonstrated superior performance compared to other models.

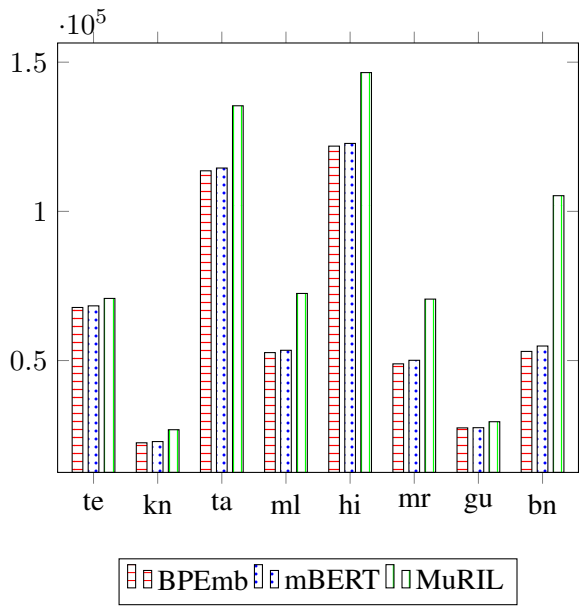
\begin {figure}[!t]
\centering
\begin{tikzpicture}
  \begin{axis}[  
    ybar, 
    bar width=4pt,
    enlargelimits=0.08,
    legend style={at={(0.5,-0.2)}, 
      anchor=north,legend columns=-1},     
    symbolic x coords={te, kn, ta, ml, hi, mr, gu,  bn},  
    xtick=data,  
    ]  
\addplot[pattern = horizontal lines, pattern color = red] coordinates {(te,67788) (kn,22437) (ta,113591) (ml,52664) (hi,121862) (mr,48895) (gu,27439) (bn,53063) }; 
\addlegendentry{BPEmb}

\addplot[pattern = dots, pattern color = blue] coordinates {(te,68331) (kn,22847) (ta,114521) (ml,53439) (hi,122765) (mr,50087) (gu,27499) (bn,54872)};
\addlegendentry{mBERT}

\addplot[pattern = vertical lines, pattern color = green] coordinates {(te,70821) (kn,26817) (ta,135385) (ml,72487) (hi,146504) (mr,70593) (gu,29500) (bn,105251)}; 
\addlegendentry{MuRIL}

  \end{axis}
\end{tikzpicture}
\caption{Language-wise count of Wikipedia articles used in training the models}
\label{fig:Wikipedia_aritcles_count}
\end{figure}

It is worth noting that the effectiveness of the models is directly linked to the amount of data on which they were trained. This is evident from Figure \ref{fig:Wikipedia_aritcles_count}, which illustrates the size of the Wikipedia data used for training BPEMb, mBERT, and MuRIL models in terms of the number of articles. Therefore, models trained using MuRIL exhibit improved performance.

Another noteworthy observation is that the Muril+LSTM model, despite being trained on a smaller dataset due to limited computational resources, performed on par with the fine-tuned MuRIL model. It is reasonable to assume that an increase in data size and/or the number of training parameters would further enhance the model's performance, resulting in improved correlation with human judgments.

\subsection{Challenges} 
One area where our models encountered challenges was in handling sentences that contained digits and abbreviations. Due to the complexity of these elements, which often carry specific meanings and context, our models struggled to evaluate the fluency of these outputs. The limited training data and the inherent complexity of digit and abbreviation recognition posed difficulties for the models to effectively capture the intended semantics.

\section{Limitations}
Given the constraints of limited computation power, we have trained our models using base configurations consisting of 12 layers. While deeper architectures are often desirable for achieving higher performance, we have made the most of the available resources to train these models effectively. Also, due to human resource limitations, we could not explore other Indic languages in creating a benchmark test set for fluency. However, we recognize the significance of expanding the scope to encompass additional Indic languages in future research endeavors.

\section{Ethical Statement}
We scrape regional news websites for source articles, under a fair usage policy. The copyright of the original news articles remains with the original authors/publishers of the articles.

\section{Conclusion and Future Work}

This paper presents our experiments on \textit{unsupervised reference-free fluency evaluation at the sentence level}. Text fluency is the ability to read words without any apparent cognitive effort and is a critical gateway to comprehension. Fluency evaluation is very important and a fundamental step to be taken for evaluating the outputs of NLG systems. Fluency evaluation of generated text further helps to improve the models or filter unacceptable generations. Despite of text fluency evaluation being a very significant task, it is often ignored. \citealt{https://doi.org/10.48550/arxiv.1809.08731} made an initiative in this direction by proposing reference-free and reference-based approaches to measure sentence-level fluency. This paper builds on their reference-free approach by investigating on various languages of Indian subcontinent.
We use \textit{SLOR} to compute the fluency scores.
We explored several models by a) training RNN-based LMs leveraging various embeddings, b) using mBERT and MuRIL for zero-shot inferences and by further c) fine-tuning mBERT and MuRIL. We performed 100+ experiments for 10 Indic languages and identified the best models. Due to the non-availability of test sets for sentence-level fluency evaluation, we collected 5K human-annotated benchmark data for 10 languages (500 samples per language). We evaluated our test results by correlating them with human judgments. Our exploration reveals that RNN-based language models trained with MuRIL embeddings stood as the winner in computing fluency scores. Also, fine-tuned MuRIL models performed better in terms of correlation with humans. It is observed that, even with lesser training data and model parameters, our model (MuRIL+LSTM) produced significant results compared to other models. Our experiments indicate that assessing the fluency of machine-generated text can be accomplished solely by the language model, without the need for references.

We believe that our pioneering work will act as a stepping stone for further research in this direction. This approach can be applied to other downstream tasks of NLG, such as Summarization, Paraphrase Generation, Translation, Dialogue Generation, Image Captioning, etc.

Further, we plan to extend our work to perform fluency evaluation at the discourse level and aim to release a larger human-annotated corpus as benchmark data for fluency evaluation.

\label{sec:conclusions}



\appendix
\label{sec:appendix}
\begin{table*}[t!]
\begin{tabular}{|l|c|c|l|}
\hline
\multicolumn{1}{|c|}{\textbf{Sentence}}                                                                                                                                                              & \textbf{\begin{tabular}[c]{@{}c@{}}Human\\ Rating\end{tabular}} & \textbf{\begin{tabular}[c]{@{}c@{}}Predicted\\ Score\end{tabular}} & \multicolumn{1}{c|}{\textbf{Description}}                                       \\ \hline
\begin{tabular}[c]{@{}l@{}}Viśākhapaṭnanlō konasāgutunna nirudyōga\\ saṅghāla āmaraṇa nirāhāra dīkṣa \\ \\ \textit{The ongoing hunger strike of the }\\\textit{ unemployed unions in Visakhapatnam.} \end{tabular}                                                                                    & 3                                                               & 3                                                                  & Perfectly fluent sentence                                                       \\ \hline
\begin{tabular}[c]{@{}l@{}}Gāyaṁ kāraṇaṅgā bhuvanēśvar kumār\\ \textbf{Aipi'el} nuṇḍi tappukunnāḍu \\ \\ \textit{Bhuvneshwar Kumar ruled out }\\ \textit{of \textbf{IPL} due to injury.} \end{tabular}                                                                     & 2                                                               & 2                                                                  & One missing word (Noun)                                                         \\ \hline
\begin{tabular}[c]{@{}l@{}}Prastuta paristhitullō prajalu kōviḍ 19\\ sūcanalu \textbf{pāṭicaḍa} tappanisari\\ ani rāṣṭra mukhyamantri telipāru \\ \\ \textit{The Chief Minister of the state said} \\ \textit{that it is mandatory for people }\\ \textit{to \textbf{follow} the instructions of }\\\textit{Covid-19 in the current situation}\end{tabular} & 2                                                               & 2                                                                  & One misspelt word (Verb)                                                        \\ \hline
\begin{tabular}[c]{@{}l@{}} Pradhāni nirṇayaṁ paṭla \\ santōṣaṅgā unna strīlu \\ varṣaṁ kāraṇaṅgā āṭa raddu \\ \\ \textit{Women are happy with PM's decision} \\
\textit{game canceled due to rain} \end{tabular}                                                                                  & 1                                                               & 1                                                                  & \begin{tabular}[c]{@{}l@{}}Coherence missing between\\ word-groups\end{tabular} \\ \hline
\begin{tabular}[c]{@{}l@{}}Mumbai iṇḍiyans pai arṣdīp siṅg 2 vikeṭs\\ tīsi man̄ci spelvēsāḍu\textbf{ ayitē kaligi yuva}\\ \textbf{pēsar unnāḍu sthiraṅgā uṇḍāli} \\ \\ \textit{Arshdeep Singh bowled a good spell }\\\textit{with 2 wickets against Mumbai Indians} \\\textit{\textbf{but a young pacer is there}}\\\textbf{\textit{should be consistent.}}\end{tabular}                                          & 1                                                               & 1                                                                  & Half of the sentence is fluent                                                  \\ \hline
\begin{tabular}[c]{@{}l@{}}\textbf{Anēka atipedda cainā prapan̄canlōn}ē\\\textbf{ sanvatsarālugā vastuvula undi} \\ \textbf{sarapharādārugā}\\ \\ Many biggest China in the world \\ for years of goods   there is as a supplier\end{tabular}                                                                        & 0                                                               & 0                                                                  & Non-fluent sentence                                                             \\ \hline
\begin{tabular}[c]{@{}l@{}}Rahīm \textbf{1950} nuṇḍi \textbf{1963}lō maraṇin̄cē \\ varaku bhārata jātīya phuṭbāl \\ jaṭṭuku mēnējargā unnāru \\ \\ \textit{Rahim was the manager of the} \\\textit{Indian national football team }\\\textit{from 1950 until his death in 1963}\end{tabular}                                                                  & 3                                                               & 2                                                                  & \begin{tabular}[c]{@{}l@{}}Failed to perform well for\\ sentences having digits\end{tabular}        \\ \hline
\begin{tabular}[c]{@{}l@{}}\textbf{Enī'ī'ār'ai cē} abhivr̥d'dhi\\  cēyabaḍina ī sākētita ī kaṣṭa \\ samayāllō mukhyamainadi\end{tabular}                                                                          & 2                                                               & 1                                                                  & \begin{tabular}[c]{@{}l@{}}Failed to perform well for\\ sentences having abbreviations\end{tabular} \\ \hline
\end{tabular}
\caption{Comparision of model output with humans for Telugu}
\label{tab:model_fluency_score}
\end{table*}

\bibliographystyle{acl_natbib}
\bibliography{custom}

\appendix

\section{Appendix}
\label{sec:appendix}

\begin{table*}[h!]
\centering
\begin{tabular}{|c|c|c|c|c|c|l}
\cline{1-6}
\textit{\begin{tabular}[c]{@{}c@{}}Language\\ Family\end{tabular}} & \multicolumn{1}{l|}{\textit{Language}} & \multicolumn{1}{l|}{Data Split} & \multicolumn{1}{l|}{\textit{\begin{tabular}[c]{@{}l@{}}Total tokens \\ (thousands)\end{tabular}}} & \multicolumn{1}{l|}{\textit{\begin{tabular}[c]{@{}l@{}}Unique tokens \\ (thousands)\end{tabular}}} & \multicolumn{1}{l|}{\textit{\begin{tabular}[c]{@{}l@{}}Avg tokens \\ per sentence\end{tabular}}} &                      \\ \cline{1-6}
\multirow{12}{*}{Dravidian}                                        & \multirow{3}{*}{Kannada}               & \textit{Train}                  & 1387                                                                                              & 125                                                                                                & 13.87                            \\ \cline{3-6}
                                                                   &                                        & \textit{Test}                   & 13.7                                                                                              & 5.7                                                                                                & 13.76                                \\ \cline{3-6}
                                                                   &                                        & \textit{Validation}             & 14                                                                                                & 5.7                                                                                                & 14.03                                 \\ \cline{2-6}
                                                                   & \multirow{3}{*}{Malayalam}             & \textit{Train}                  & 1173                                                                                              & 233                                                                                                & 11.73                                 \\ \cline{3-6}
                                                                   &                                        & \textit{Test}                   & 11.7                                                                                              & 7                                                                                                  & 11.76                                  \\ \cline{3-6}
                                                                   &                                        & \textit{Validation}             & 12                                                                                                & 7                                                                                                  & 11.97                                   \\ \cline{2-6}
                                                                   & \multirow{3}{*}{Tamil}                 & \textit{Train}                  & 1348                                                                                              & 127                                                                                                & 13.49                                    \\ \cline{3-6}
                                                                   &                                        & \textit{Test}                   & 13.6                                                                                              & 6.3                                                                                                & 13.65                                    \\ \cline{3-6}
                                                                   &                                        & \textit{Validation}             & 13                                                                                                & 5.9                                                                                                & 13.45                                    \\ \cline{2-6}
                                                                   & \multirow{3}{*}{Telugu}                & \textit{Train}                  & 1224                                                                                              & 142                                                                                                & 12.24                                  \\ \cline{3-6}
                                                                   &                                        & \textit{Test}                   & \textit{12}                                                                                       & \textit{6}                                                                                         & 12.24                                 \\ \cline{3-6}
                                                                   &                                        & \textit{Validation}             & \textit{12}                                                                                       & \textit{6}                                                                                         & 12.2                                 \\ \cline{1-6} \cline{1-6} \cline{1-6} \cline{1-6} \cline{1-6}
\multirow{18}{*}{Indo-Aryan}                                       & \multirow{3}{*}{Bengali}               & \textit{Train}                  & 1308                                                                                              & 69                                                                                                 & 12.96                                                                                            &                      \\ \cline{3-6}
                                                                   &                                        & \textit{Test}                   & 13                                                                                                & 4.5                                                                                                & 13.02                                                                                            &                      \\ \cline{3-6}
                                                                   &                                        & \textit{Validation}             & 13                                                                                                & 4.5                                                                                                & 12.93                                                                                            &                      \\ \cline{2-6}
                                                                   & \multirow{3}{*}{Gujarati}              & \textit{Train}                  & 1461                                                                                              & 128                                                                                                & 14.61                                                                                            &                      \\ \cline{3-6}
                                                                   &                                        & \textit{Test}                   & 14                                                                                                & 5.3                                                                                                & 14.43                                                                                            &                      \\ \cline{3-6}
                                                                   &                                        & \textit{Validation}             & 14.7                                                                                              & 5.3                                                                                                & 14.8                                                                                             &                      \\ \cline{2-6}
                                                                   & \multirow{3}{*}{Hindi}                 & \textit{Train}                  & 1701                                                                                              & 60                                                                                                 & 17.01                                                                                            &                      \\ \cline{3-6}
                                                                   &                                        & \textit{Test}                   & 16.7                                                                                              & 4                                                                                                  & 16.72                                                                                            &                      \\ \cline{3-6}
                                                                   &                                        & \textit{Validation}             & 16.8                                                                                              & 3.8                                                                                                & 16.85                                                                                            &                      \\ \cline{2-6}
                                                                   & \multirow{3}{*}{Marathi}               & \textit{Train}                  & 1353                                                                                              & 104                                                                                                & 13.53                                                                                            &                      \\ \cline{3-6}
                                                                   &                                        & \textit{Test}                   & 13                                                                                                & 5                                                                                                  & 13.27                                                                                            &                      \\ \cline{3-6}
                                                                   &                                        & \textit{Validation}             & 13.6                                                                                              & 4.8                                                                                                & 13.69                                                                                            &                      \\ \cline{2-6}
                                                                   & \multirow{3}{*}{Odia}                 & \textit{Train}                  & 1427                                                                                              & 80                                                                                                 & 13.37                                                                                            &                      \\ \cline{3-6}
                                                                   &                                        & \textit{Test}                   & 14                                                                                                & 4.5                                                                                                & 13.11                                                                                            &                      \\ \cline{3-6}
                                                                   &                                        & \textit{Validation}             & 14.5                                                                                              & 4.5                                                                                                & 13.47                                                                                            &                      \\ \cline{2-6}
                                                                   & \multirow{3}{*}{Sinhala}               & \textit{Train}                  & 1591                                                                                              & 75                                                                                                 & 16.31                                                                                            &                      \\ \cline{3-6}
                                                                   &                                        & \textit{Test}                   & 16                                                                                                & 5                                                                                                  & 16.29                                                                                            &                      \\ \cline{3-6}
                                                                   &                                        & \textit{Validation}             & 16                                                                                                & 5                                                                                                  & 16.42                                                                                            &                      \\ \cline{1-6}
\end{tabular}
\caption{Data Statistics}
\label{tab:data_statistics}
\end{table*}

\begin{table*}[h!]
\centering
\begin{tabular}{|c|c|c|}
\hline
\textbf{SNo} & \textbf{Language} & \textbf{Website}                 \\ \hline
1            & Telugu       & \url{https://www.vaartha.com/}         \\ \hline
2            & Hindi        & \url{https://www.indiatv.in/}          \\ \hline
3            & Tamil        & \url{https://www.updatenews360.com/}   \\ \hline
4            & Kannada      & \url{https://kannadanewsnow.com/kannada/}         \\ \hline
5            & Malayalam    & \url{https://dailyindianherald.com/}   \\ \hline
6            & Bengali      & \url{https://www.abplive.com/}         \\ \hline
7            & Gujarati    & \url{https://www.gujaratsamachar.com/} \\ \hline
8            & Marathi      & \url{https://www.abplive.com/}         \\ \hline
9            & Odia      & \url{https://www.dharitri.com/}         \\ \hline
10            & Sinhala      & \url{https://www.newsfirst.lk/sinhala/}         \\ \hline
\end{tabular}
\caption{News articles crawling sources}
\label{tab:crawling_sources}
\end{table*}

\begin{table*}[h!]
\centering
\begin{tabular}{|l|c|c|c|}
\hline
\textbf{}                    & \multicolumn{1}{l|}{\textbf{LSTM + Muril}} & \multicolumn{1}{l|}{\textbf{Muril w/ finetuning}} & \multicolumn{1}{l|}{\textbf{mBERT w/ finetuning}} \\ \hline
\textit{Sequence Length}     & 128                                        & 128                                               & 128                                               \\ \hline
\textit{Embedding dimension} & 768                                        & 768                                               & 768                                               \\ \hline
\textit{Learning Rate}       & 0.001                                      & $2e^{-5}$                                              & $2e^{-5}$                                              \\ \hline
\textit{Batch size}          & 256                                        & 8                                                 & 8                                                 \\ \hline
\textit{Epochs}              & 5                                          & 5                                                 & 5                                                 \\ \hline
\textit{hidden units}        & 512                                        & -                                                 & -                                                 \\ \hline
\textit{\# Layers}           & 2                                          & 12                                                & 12                                                \\ \hline
\end{tabular}
\caption{Model Hyper Parameters}
\label{tab:model_hyperparameters}
\end{table*}

\end{document}